\documentclass[sigconf]{acmart}

\AtBeginDocument{%
  \providecommand\BibTeX{{%
    \normalfont B\kern-0.5em{\scshape i\kern-0.25em b}\kern-0.8em\TeX}}}

\acmSubmissionID{rtp2615}

\usepackage{amsmath}
\usepackage{graphicx}
\usepackage{booktabs}
\usepackage{wrapfig}

\copyrightyear{2024}
\acmYear{2024}
\setcopyright{rightsretained}
\acmConference[KDD '24]{Proceedings of the 30th ACM SIGKDD Conference on
Knowledge Discovery and Data Mining}{August 25--29, 2024}{Barcelona, Spain}
\acmBooktitle{Proceedings of the 30th ACM SIGKDD Conference on Knowledge
Discovery and Data Mining (KDD '24), August 25--29, 2024, Barcelona,
Spain}\acmDOI{10.1145/3637528.3672062}
\acmISBN{979-8-4007-0490-1/24/08}

\begin{document}
\title{RHiOTS: A Framework for Evaluating Hierarchical Time Series Forecasting Algorithms}

\author{Luis Roque}
\email{luis_roque@live.com}
\orcid{0000-0002-0899-3209}
\affiliation{%
  \institution{LIACC/Faculty of Engineering, University of Porto}
  \city{Porto}
  \country{Portugal}
}

\author{Carlos Soares}
\affiliation{%
  \institution{LIACC/Faculty of Engineering, University of Porto}
  \city{Porto}
  \country{Portugal}
}
\affiliation{%
  \institution{Fraunhofer AICOS Portugal}
  \city{Porto}
  \country{Portugal}
}

\author{Luís Torgo}
\affiliation{%
  \institution{Dalhousie University}
  \city{Halifax}
  \country{Canada}
}

\renewcommand{\shortauthors}{Luis Roque, Carlos Soares, and Luís Torgo}

\begin{abstract}
We introduce the Robustness of Hierarchically Organized Time Series (RHiOTS) framework, designed to assess the robustness of hierarchical time series forecasting models and algorithms on real-world datasets. Hierarchical time series, where lower-level forecasts must sum to upper-level ones, are prevalent in various contexts, such as retail sales across countries. Current empirical evaluations of forecasting methods are often limited to a small set of benchmark datasets, offering a narrow view of algorithm behavior. RHiOTS addresses this gap by systematically altering existing datasets and modifying the characteristics of individual series and their interrelations. It uses a set of parameterizable transformations to simulate those changes in the data distribution.
Additionally, RHiOTS incorporates an innovative visualization component, turning complex, multidimensional robustness evaluation results into intuitive, easily interpretable visuals. This approach allows an in-depth analysis of algorithm and model behavior under diverse conditions. 
We illustrate the use of RHiOTS by analyzing the predictive performance of several algorithms. Our findings show that traditional statistical methods are more robust than state-of-the-art deep learning algorithms, except when the transformation effect is highly disruptive. Furthermore, we found no significant differences in the robustness of the algorithms when applying specific reconciliation methods, such as MinT.
RHiOTS provides researchers with a comprehensive tool for understanding the nuanced behavior of forecasting algorithms, offering a more reliable basis for selecting the most appropriate method for a given problem.
\end{abstract}



\maketitle

\section{Introduction}


In time series forecasting, modeling inter-temporal dependencies between observations is a crucial task to capture the dynamics of the underlying process. Furthermore, considering cross-series information, i.e., the relationship between multiple related time series, can enhance the performance of traditional univariate algorithms~\cite{hyndman2018optimal}. Such relationships occur in many real-world situations and often represent hierarchical structures, such as in data concerning Gross Domestic Product~\cite{Athanasopoulos2020}, epidemics~\cite{gitto2021forecasting}, and sales demand~\cite{KARMY201959}. Coherent forecasts~\cite{hyndman2018optimal}, which satisfy these hierarchical relations, are often required in such applications. To improve performance, state-of-the-art algorithms for hierarchical time series (HTS) forecasting rely on both the autocorrelation of each time series and the cross-series correlations~\cite{hyndman2018forecasting,hts_elec_hyndman}.



An important characteristic of HTS forecasting algorithms is their robustness to changes in time series relationships, such as variations in seasonality, trends, cross-series dependencies, and volatility. These phenomena significantly impact predictive performance and may lead to inaccurate forecasts and suboptimal decision-making. Nevertheless, conventional evaluation methodologies, typically based on a small number of datasets and metrics, cannot assess the robustness of HTS forecasting algorithms in dynamic scenarios where relationships between time series and temporal dependencies change over time. We note that the concerns about the adequacy of evaluation methods are not limited to HTS forecasting. In fact, there is a growing discussion on the limitations of empirical evaluations in time series analysis (e.g., anomaly detection~\cite{Keogh_anomaly_det}). We argue that similar issues can be raised for time series forecasting since both suffer the exact root cause: most papers evaluate algorithms on one or more of a handful of popular benchmark datasets (e.g.~\cite{UCI_PEMS_SF,UCI_electricity}).




This work introduces the Robustness of Hierarchically Organized Time Series (RHiOTS) framework. RHiOTS is designed to evaluate the robustness of HTS forecasting models and algorithms facing real-world data distribution changes. 
Our framework systematically quantifies the robustness by imposing controlled realistic transformations, mimicking common patterns observed in actual datasets. These semi-synthetic datasets retain essential characteristics of the original data while introducing new dynamics, serving as a robust baseline to assess algorithmic performance against data variations.
RHiOTS has an innovative visualization component, which transforms the complex, multidimensional results of the robustness evaluation into intuitive, easily interpretable visual representations. It focuses on aggregating the main findings on the performance of different algorithms and understanding the impact of the various data transformations. It enables practitioners to make more informed data-driven decisions in selecting and applying HTS forecasting algorithms.


To illustrate the usefulness of RHiOTS, we evaluated five forecasting algorithms across three real-world time series datasets. Our approach involved creating variations of these datasets using four time series transformations, often used in the context of time series augmentation. For each transformation, we defined a set of parameter variants and then generated several samples of these transformed datasets to reduce the variability of the results. 

The results indicate that traditional statistical algorithms, such as Exponential Smoothing (ETS), show considerable robustness and ranked best using the RHiOTS framework. Also, we found no meaningful differences in the robustness of the algorithms when applying specific reconciliation methods, such as MinT. An additional observation is that, while generally less consistent, deep learning counterparts showed more robustness when dealing with disruptive transformations, such as high-intensity magnitude warping. We also demonstrated that these insights could not have been obtained through the limited benchmark-based methodologies currently in use.

%
In summary, the contributions of this paper are the following:
\begin{itemize}
\item We introduce RHiOTS, a novel framework for evaluating the robustness of HTS forecasting models and algorithms on real-world datasets. It provides a more systematic and comprehensive comparison than the existing benchmark-based methods;
\item We propose a visual representation of the results of RHiOTS. It converts complex multidimensional results into clear, concise plots, enabling easier comparison of performance and the effects of data transformations;
\item We present empirical insights into the robustness of five distinct HTS models and algorithms. It illustrates the usefulness of RHiOTS as a new evaluation framework.
\end{itemize}

All experiments are fully reproducible, and the methods and time series data are available in a public code repository.\protect\footnotemark{}

\footnotetext{\url{https://github.com/luisroque}\\\url{/robustness_hierarchical_time_series_forecasting_algorithms}}

\section{Background and Notation}
\label{chap:background}

We are working with a collection of $S$ related univariate time series, $\mathcal{Z} = \{\textbf{z}^i_t, t \in \mathbb{N}, i = 1, \dots, S\}$. The training values can be written as $\textbf{z}^i_{1:T} = [z^i_1, z^i_2, \cdots, z^i_T]$ where $z^i_t \in \mathbb{R}$ denotes the value of time series $i$ at time $t$ and $T$ represents the last training point. When the interpretation is unambiguous and to simplify the notation in specific sections, we refer to $\textbf{z}^i$ as the observed time series. The training range is denoted by $\{1,2,...,T\}$, while $\{T+ 1,T + 2, \dots, T+\tau\}$ is the prediction range and $\tau$ is the forecast horizon. Point predictions are defined as $\hat{\textbf{z}}^i_{T+1:T+\tau}$.

\begin{figure}[t]
\centering
\includegraphics[height=5.5cm]{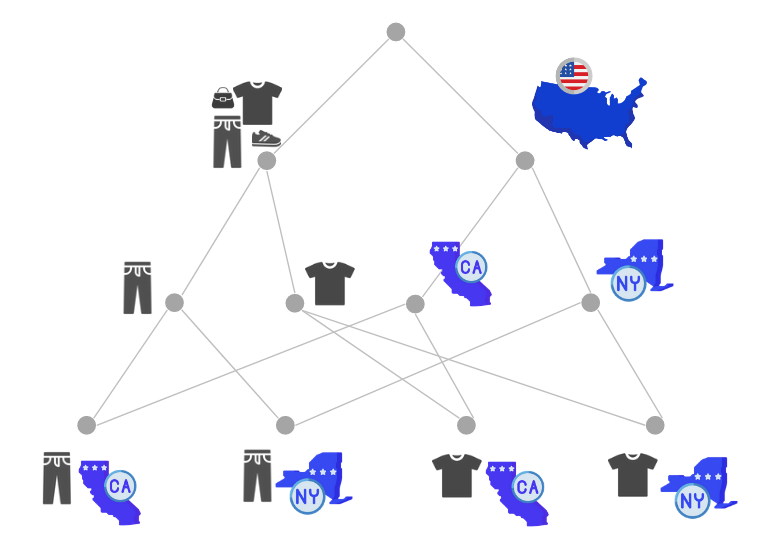}
\Description{This is a simple example of a hierarchically organized time series dataset that comprises sales data from a retailer in the US. The figure shows different levels of sales data aggregation.}
\caption{Simple example of a hierarchically organized time series dataset that comprises sales data from a retailer in the US.}
\label{fig:group_diag}
\end{figure}

HTS organizes a collection of time series, represented as $\textbf{z}^i$, within a hierarchical structure denoted by $H$. The hierarchy $H$ is a tree-like structure, where each node represents a group $G_l$ or a time series $\textbf{z}^i$. The root of the tree represents the total aggregate of all the time series in the dataset, denoted as $\textbf{z}_t$. The leaf nodes of the tree represent individual time series $\textbf{z}^i$. Each time series $\textbf{z}^i$ is associated with one or more groups $G_l$ in the hierarchy. Thus, we can write $G_l \subseteq {1,...,S}$ where $\textbf{z}^{G_l} = \sum_{i \in G_{l}} \textbf{z}^i$. The subset of groups a time series $\textbf{z}^i$ belongs to is denoted by $L_i = \{l: i \in G_l\}$. $G$ denotes the set of all groups, and its cardinality is denoted by $|G|$.

We illustrate this concept in Figure~\ref{fig:group_diag} using an example from a retailer operating in the United States. It consists of two groups: The first, denoted by $U$, represents the states in the United States (US), with elements $a$ and $b$, representing California (CA) and New York (NY), respectively. The second group, denoted by $P$, represents product categories and has elements $x$ and $y$, corresponding to trousers and t-shirts, respectively. At the bottom level, the hierarchy would generate four distinct time series: $\textbf{z}^{ax}_{t}$, $\textbf{z}^{ay}_{t}$, $\textbf{z}^{bx}_{t}$, and $\textbf{z}^{by}_{t}$. Each non-leaf node (group $G_l$) in the hierarchy represents the aggregation of the time series associated with its children nodes. For instance, if $U$ represents the group of all stores in different states of the US, and $\textbf{z}^i$ represents the sales data of an individual product in a specific store, then the time series at node $U$ is $\textbf{z}_{U,t} = \sum_{\textbf{z}^i \in U} \textbf{z}^i_t$. Similarly, the time series for each element $a$ of $U$ is $\textbf{z}^{a}_{t} = \sum_{i \in U_{a}} \textbf{z}^i_t$.

The objective of HTS forecasting is twofold: firstly, to minimize the forecast error for each series within the hierarchy, and secondly, to ensure that forecasts at all levels of the hierarchy are consistent with each other. Formally, the forecast error for a series $i$ over a prediction horizon $\tau$ is defined as $E^i_{T+1:T+\tau} = \{ e^i_{t} = \hat{z}^i_{t} - z^i_{t} \,|\, t \in \{T+1, T+2, \ldots, T+\tau\} \}$, where $\hat{z}^i_{t}$ denotes the forecasted value and $z^i_{t}$ denotes the actual value at time $t$. The consistency constraint requires that for any aggregated level within the hierarchy, represented by a group $G_l$, the aggregated forecasts $\hat{\textbf{z}}^{G_l}_{T+1:T+\tau} = \sum_{i \in G_l} \hat{\textbf{z}}^i_{T+1:T+\tau}$, should equal the sum of forecasted values for all time series $i$ belonging to that group. This ensures that the forecasts respect the hierarchical structure, maintaining integrity and coherence across all levels of aggregation within the dataset.

In terms of evaluation and using our previous example, we start by computing the Mean Absolute Scaled Error (MASE)~(\cite{fpp3}) for each time series at the bottom level, such as $MASE_{ax}$ for the time series $\textbf{z}^{ax}$. The formula for MASE is defined as follows: $MASE_{i} = \frac{1}{\tau - T + 1}\sum_{t=T+1}^{\tau}\frac{|{e}^i_t|}{\frac{1}{(T-1)}\sum_{t=2}^{T}|{z}^i_{t}-{z}^i_{t-1}|}$.

It is equally important to assess the forecasting performance at aggregated levels. Hence, we also compute the MASE metric for every group element $l$. This is achieved by aggregating the time series at the bottom level that belongs to a specific group element. For example, for group element $a$ in our previous example, the aggregated time series $\textbf{z}^{a}_{t}$ is calculated as $\textbf{z}^{a}_{t} = \textbf{z}^{ax}_{t} + \textbf{z}^{ay}_{t}$. Subsequently, the MASE metric for $\textbf{z}^{a}_{t}$ is computed.

To evaluate the forecast accuracy at the group level, we calculate the average MASE across all elements $l$ that belong to the group. For example, for group $U$, we would consider $\textbf{z}^{a}_{t}$ and $\textbf{z}^{b}_{t}$. First, we define the aggregated time series for each group $l$ as 
$\textbf{z}^{G_l} = \sum_{i:i \in G_l}\textbf{z}^i$.

Then, we compute the MASE for each of these aggregate time series $\textbf{z}^{G_l}$ and finally take the average across all $l$ in the group $G$, $MASE_{G} = \frac{1}{|G|}\sum_{l \in G} MASE_{G_l}$.

At the most aggregated level (the top of the hierarchy), we sum the values of all the bottom-level time series and evaluate the predictive performance based on these values. 

Analyzing the distance between time series is not a trivial task. We adopt Dynamic Time Warping (DTW) following the approach recommended by~\cite{Keogh2004OnTN} for short time series. DTW evaluates sequence similarity by computing a matrix of distances between elements and determining the optimal alignment path that minimizes the total distance (Eq.~\ref{eq:dtw}). This alignment, or warp path, is mathematically formulated as:

\begin{eqnarray}
    \label{eq:dtw}
    DTW_q(\textbf{z}, \textbf{z}')=\min_{\pi \in A(\textbf{z}, \textbf{z}')}\bigg( \sum_{(i,j) \in \pi} d(\textbf{z}_i, \textbf{z}_j')^q \bigg)^\frac{1}{q}
\end{eqnarray}
where an alignment path $\pi$ of length $K$ is a sequence of $K$ index pairs $((i_0,j_0),...,(i_{K-1}, j_{K-1}))$ and $A(\textbf{z},\textbf{z}')$ is the set of all admissible paths.

In terms of time series transformations, we denote them by the function $\mathcal{T}$, to the original time series $\textbf{z}^{i}_{t}$. Such transformations can modify individual components of a time series and alter the interrelations among multiple time series within the dataset. Different transformation functions will have different sets of parameters. For simplicity of notation, we introduce the concept of a parameter set $v$, representing a specific parameter set used in the transformation. Formally, we denote the resulting transformed series as $\textbf{z}^{'i}_{v, t}$, where $\textbf{z}^{'i}_{v, t} = \mathcal{T}(\textbf{z}^{i}_{t}, \eta_v)$, and $\eta_v$ symbolizes the particular parameter set $v$. We denote the resulting dataset as $\mathcal{Z}_{\gamma,v,j}$, where $\gamma \in \{1, 2, \ldots, \Gamma\}$ denotes the transformation applied, $v$ is the set of parameters used, and $j$ represents the number of samples generated from the same transformation and set of parameters.

\section{Related Work}
\label{chap:RW}

\begin{figure*}[t]
  \centering
  \includegraphics[height=4.5cm]{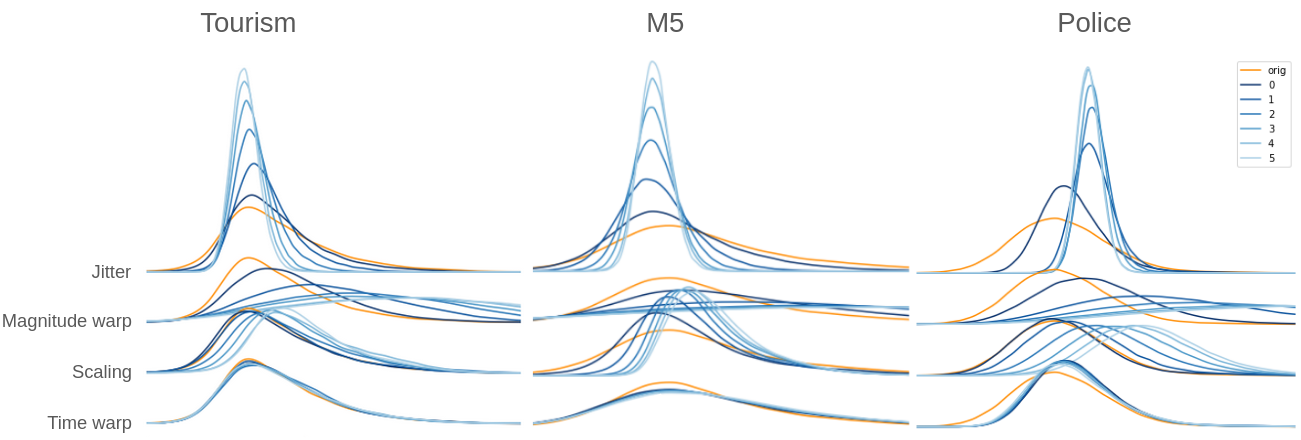}
  \Description{A ridge plot showing the Dynamic Time Warping (DTW) distribution between time series in the dataset. The plot is divided into columns representing datasets and rows representing transformations and parameter sets. The original DTW distances are shown in orange, while the transformed distances are shown in shades of blue. As the magnitude of the transformation increases, the blue color becomes lighter.}
  \caption{Ridge plot that shows the DTW distribution between time series in the dataset for each dataset (columns), transformation (rows), and parameter set. For the original DTW distance we use orange and for the transformed ones we use shades of blue:  as we increase the magnitude of the transformation the color gets lighter.}
  \label{fig:ridge}
\end{figure*}

\textbf{Hierarchical Time Series Forecasting Methods.} Univariate methods like ARIMA and ETS are common for individual time series analysis and can be used for HTS when using simple reconciliation strategies~\cite{fpp3}. General approaches like those found in~\cite{smyl,deepAR_prob,nbeats,lim2020temporal} are methods used to forecast a set of time series from the same domain. It means that they are capable of learning relationships among time series. However, these approaches do not incorporate any hierarchical structure, missing out on leveraging it to improve forecasting accuracy.

Alternatively, several methods explicitly model the hierarchical nature of HTS datasets. Early approaches by~\cite{hyndman2011optimal}, followed by refinements in~\cite{ATHANASOPOULOS2009146,hyndman2018optimal}, involve fitting and reconciling independent forecasts at all hierarchy levels. Non-linear models (e.g.,~\cite{sharq}) improve their ability to capture complex patterns in the data by employing optimization and regularization techniques during the model training process. Another approach uses Gaussian Processes and does not require any reconciliation since the hierarchical structure is an input to the model itself~\cite{gphf}.

\textbf{Evaluating Hierarchical Time Series Forecasting}. 
The evaluation of HTS forecasting models has received limited attention in recent literature. Exceptions. For example, \cite{sharq} and~\cite{hierar_athanasopoulos}, compare the performance of state-of-the-art reconciliation models on a limited set of real-world datasets. Despite considering various forecasting models, the fact that the experimental setup does is so limited and does not provide rich data variations, our ability to generalize is limited. Recently, a practical guide addressing concepts like scale, units, sparsity, forecast horizon, multiple evaluation windows, and decision context has been proposed~\cite{EvaluationHierarchicalForecasts}. However, the guide falls short in objectively quantifying performance differences between models beyond basic accuracy assessment on benchmarks.

\textbf{Time Series Augmentation.} Data augmentation involves generating synthetic data that covers unexplored input space while preserving correct labels~\cite{Wen_2021}. This technique has proven effective in domains such as computer vision, where methods like AlexNet~\cite{imagenet} have leveraged augmented data for image classification. On the other side, the unique properties of time series data, such as its temporal dependencies and intricate dynamics, create a set of challenges. 

Autoregressive models, while effective for linear, stationary time series, face limitations with complex data and computational demands~\cite{gratis}. Generative models, including Generative Adversarial Networks (GANs) and Variational Autoencoders (VAEs), have emerged as powerful tools for augmenting time series data. GANs, particularly TimeGAN, and extensions of this work, create realistic synthetic samples by learning temporal dynamics, albeit with training challenges~\cite{Wen_2021,timegan,ni2021sigwasserstein}. VAEs and their conditional variants (CVAEs) generate new data from a latent space, offering potential despite some control issues over the generated samples~\cite{Wen_2021,vae_aug,vae_ts,vrnn}.

Sliding window methods risk overfitting by focusing too narrowly on local patterns, while decomposition methods generate limited variety from extracted dataset features like trends~\cite{decomp_aug}.

Simpler transformations like jittering, scaling, magnitude, and time warping have been shown to help in specific tasks and domains~\cite{jitter,scaling_aug}. They are simple to implement and increase data variety. While they may not capture complex patterns, they provide some control over the transformations since they are parametric transformations.

\section{RHiOTS}
\label{chap:model}

We introduce RHiOTS, a framework designed to assess the robustness of HTS forecasting algorithms. Traditional evaluations, which primarily focus on predictive accuracy using a limited set of benchmark problems, fail to adequately assess the resilience of an algorithm to minor variations in individual time series or their interrelations. RHiOTS addresses these issues by offering a nuanced analysis of the stability of forecasting algorithms against such changes, providing deeper insights into their robustness and reliability.

\subsection{Framework}
\label{chap:framework}

RHiOTS serves two primary objectives. The first goal is to provide a comprehensive understanding of the behavior of models. We do this by systematically applying various transformations to the data and then evaluating the performance of models. This process facilitates a detailed assessment of the robustness of each model, controlled by transformation and intensity. It enables practitioners to identify the most effective model for the unique aspects of their problem and potential variations in data properties.

For the second goal, RHiOTS aims to support the reliable selection of an appropriate algorithm for a given forecasting task. By analyzing how different algorithms perform across a range of transformed datasets, the framework offers insights into which algorithms are most adaptable and effective under varying conditions. This allows for a more effective and informed comparison of algorithms and improves generalization.


RHiOTS applies transformations to each individual time series, $\textbf{z}^{'i} = \mathcal{T}^{i}(\textbf{z}^{i}, \eta^{i})$. The transformations are applied to the time series in the leaf nodes of the hierarchy only. The aggregated levels of the hierarchy are recomputed after the transformation, i.e., the total sum of the observations of a specific group (or for the top-level series) is computed based on the transformed individual series, $\textbf{z'}^{G} = \sum_{l:\textbf{z'}^i \in G_l} \textbf{z'}^{G_l}$.

The next step in RHiOTS is to assess the robustness of HTS forecasting models by connecting the variations introduced by time series transformations with the variation in forecasting performance. To quantify the variation in forecasting performance, we compute the forecasting error of the model on the original dataset and the various transformed versions. Given that each transformation represents a type of variation (e.g., jittering represents noise in the measurement of the values), the analysis of the variation of forecasting performance for different versions of a transformation gives a systematic perspective on the robustness of a method to that type of variation (e.g., the robustness of the method to noise).

\begin{figure*}[t]
\centering
\includegraphics[width=0.85\textwidth]{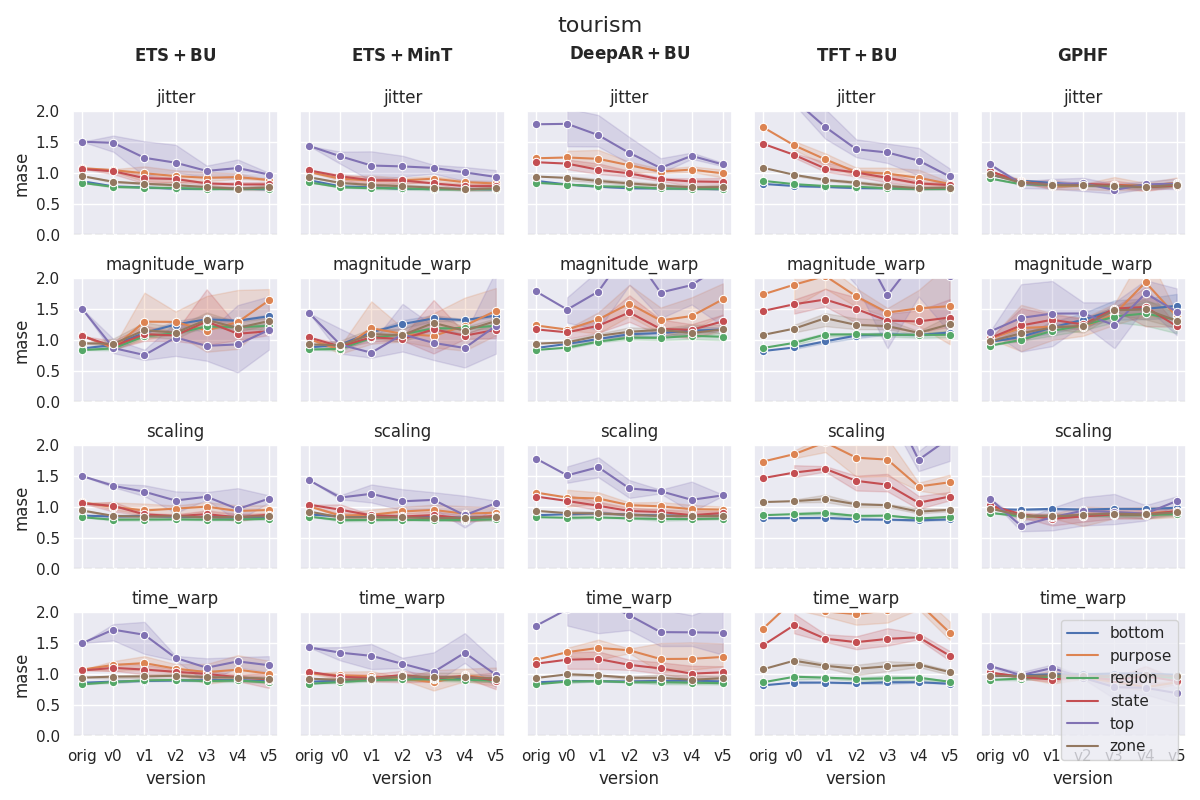}
\Description{A series of panels showing model performance across various data transformations in hierarchical time series forecasting, assessed using MASE. Each panel represents a different forecasting model subjected to transformations such as jitter, magnitude warp, scaling, and time warp, with the transformation intensity increasing from 'orig' to 'v5'. Lines within each panel correspond to different hierarchical levels of the data, providing insight into the robustness of each model at various granularities.}
\caption{Model performance across various data transformations in hierarchical time series forecasting, assessed using MASE. Each panel represents a different forecasting model subjected to transformations such as jitter, magnitude warping, scaling, and time warping, with the transformation intensity increasing from `orig` to `v5`. The lines within each panel correspond to different hierarchical levels of the data, providing insight into the robustness of each model at various granularities.}
\label{fig:results}
\end{figure*}

\subsection{Time Series Transformations}
\label{chap:tranfs}

We apply random-based transformations to the original time series, represented by $\textbf{z}^{'i}_{v, t} = \mathcal{T}(\textbf{z}^{i}_{v, t}, \eta)$, where $\mathcal{T}$ denotes a transformation function and $\eta$ indicates its governing parameters. These transformations can affect individual time series components and relationships between series in a dataset. However, they should be smooth and continuous to preserve a meaningful relationship between the original and transformed series. This ensures that similar parameters produce closely related transformed series.


Jittering is a magnitude domain transformation that can be defined as the addition of a random noise component to the values of a time series $z_{v,t}^{'i} = z_{t}^i + \epsilon^i_{v,t}$ where $\epsilon^i_{v,t} \sim \mathcal{N}(0, \sigma_v^2)$.
the standard deviation $\sigma_{v, i}$ of the added noise is a transformation parameter. Applying jittering to the time series using the addition of i.i.d. Gaussian noise is widely used in the literature to simulate realistic sources of measurement error or irregularity in time series data~\cite{Iwana_2021}. For example, consider a retail store where
sales data suddenly becomes more erratic due to unexpected
external factors such as local construction affecting customer
traffic.

Another transformation used is scaling, which involves modifying the amplitude of the series by a random scalar value ${z}_{v,t}^{'i} = \alpha^i_{v} {z}_{t}^i$ where $\alpha^i_{v} \sim \mathcal{N}(0, \sigma_{v,i}^2)$. Once again, $\sigma_{v, i}$ defines the standard deviation of the multiplicative effect for each parameter set of the transformation, dataset, and series. Scaling the time series data using a multiplicative factor simulates realistic changes in the magnitude of the time series. For example, consider a retail store experiencing increased variability in sales due to various promotional campaigns executed by the store itself and its competitors

We also applied magnitude warping~\cite{Iwana_2021}. This method causes a smooth, continuous, nonlinear transformation of time series data. It can be written as ${z}^{'i}_{v, t} = S^i_{{u}^i_k}({z}^{i}_{v, t})$ where ${u}^i_k \sim \mathcal{N}(1, \sigma_{v,i}^2)$. Note that $S^i_{{u}^i_k}$ interpolates a cubic spline with knots $\textbf{u}=u_{1}, ..., u_{k}$. Each knot $u_k$ comes from a distribution $\mathcal{N}(1, \sigma_{v, i}^2)$, with the number of knots $k$ and the standard deviation $\sigma_{v, i}$ as parameters. The cubic spline is fitted to the original data points, and the transformed data is obtained by scaling the original magnitude using the evaluated cubic spline function values. The proposed transformation method modulates the magnitude of the time series, maintaining its overall structure and smoothness. For example, the summer season sales time series of a store near a beach area during unusually high temperatures could experience magnitude warping.

Finally, the last transformation considered was time warping. Time warping is a process that stretches or compresses the time axis of a time series. Similarly to magnitude warping, we define time warping using a cubic spline to interpolate the values of the time series at a set of evenly spaced time points. The time points can be chosen such that they are spaced more closely together in regions where the time series is stretched and more widely spaced in regions where it is compressed. This will effectively stretch or compress the time axis of the time series while preserving the shape of the time series itself. Time warping is defined as ${z}^{'i}_{v, t} = {z}^{i}_{v, S_{{u}^i_k}(t)}$ where ${u}^i_k \sim \mathcal{N}(1, \sigma_{v,i}^2)$.
Now, $S^i_{{u}^i_k}(t)$ is applied on the time steps. Time warping can potentially impact the seasonality of a time series by stretching or compressing the time axis of the series. This can cause the periodic patterns in the series, such as seasonal cycles or trends, to become more or less prominent, depending on how the time axis is altered. For example, changes in weather cycles could significantly impact the seasonality of specific stores and products.

\section{Experiments}
\label{chap:results}

The primary objective of our experimental setup is to illustrate how RHiOTS can be used to analyze the robustness of both models and algorithms within the context of Hierarchical Time Series (HTS) forecasting.

The first step we are interested in is how different transformations affect the distance between time series in the dataset (\textbf{Q1}). HTS algorithms rely on these dependencies to improve their univariate estimates.

\begin{enumerate}
    \item \textbf{Models:} In selecting models for a specific problem, standard evaluation methods test those models on the available data. The assumption is that existing data is representative of new, unseen data. RHiOTS can be used to test the robustness of those models to variations in the data (e.g., noise). We investigate the effects of different types of perturbations on prediction error (\textbf{Q2}) and how the prediction error of HTS forecasting models changes with manipulation of dependencies across time and between series (\textbf{Q3}). Then, we use RHiOTS to systematically compare the impact of the data transformations on ranking the different algorithms (\textbf{Q4}). This kind of analysis could support the data scientist in choosing the best model not only in terms of standard evaluation methods (e.g., average forecasting performance) but also in terms of their robustness (e.g., robustness of forecasting performance to noise in the data).

    \item \textbf{Algorithms:} In developing new algorithms, typical benchmarking methods test those algorithms on public datasets available for research. The results of the new algorithm are compared to the results obtained by simple baselines and state-of-the-art algorithms (e.g., the average forecasting accuracy). Thus, we start by comparing the performance of the algorithms using the benchmark method of assessing predictive performance (\textbf{Q5}). Then, we use RHiOTS to assess the relevance of dependencies between time series for algorithm performance (\textbf{Q6}) and to identify the most robust algorithm suitable for a given application domain (\textbf{Q7}). This evaluation process enables researchers to make informed decisions when developing new HTS algorithms and dealing with diverse datasets and application requirements.
\end{enumerate}








\subsection{Datasets}
The empirical evaluation uses three public datasets: the Tourism~\cite{tourism} dataset from the Australian Bureau of Statistics, the M5~\cite{MAKRIDAKIS2021} dataset based on Walmart sales, and the Police~\cite{houston} dataset from Houston police criminal reports. These datasets encompass various time granularities, frequencies, lengths, and hierarchical structures, illustrating the usefulness of RHiOTS in evaluating the robustness of time series forecasting models in different scenarios.

To efficiently conduct the experiments across all datasets and models, we downsampled the M5 dataset by reducing its frequency from daily to weekly. Additionally, for the M5 and Houston datasets, we selected a subset of 500 time series with high count levels. Preliminary experiments indicate that this selection has no significant impact on evaluating the different algorithms using RHiOTS. In the interest of space, we do not discuss them here.


\subsection{Transformations and Algorithms}

\begin{figure}[t]
\centering
\includegraphics[width=0.47\textwidth]{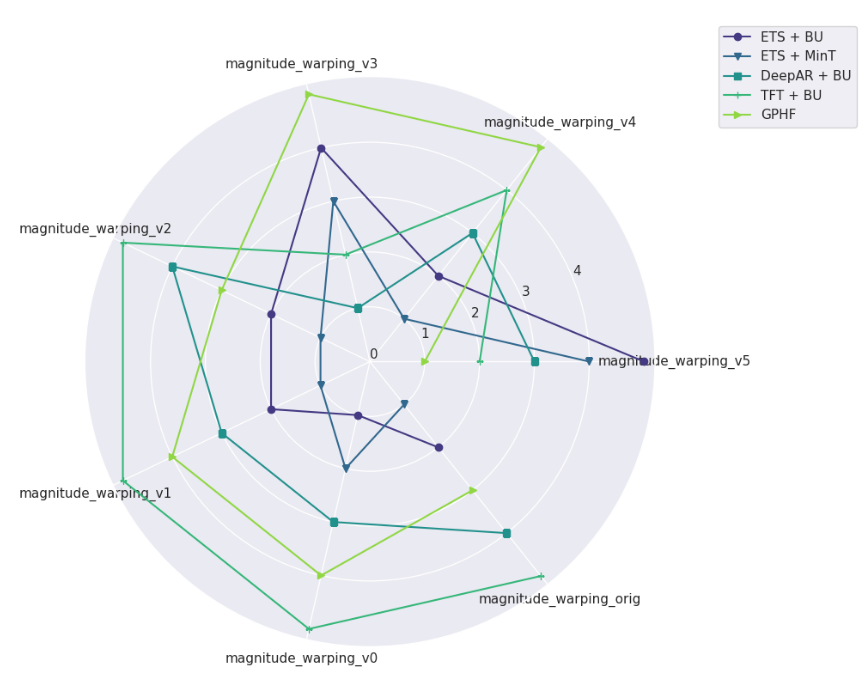}
\Description{A ranking chart showing the performance of forecasting methods under magnitude warping transformation for the Tourism dataset. The chart illustrates performance from the original data ('orig') to the most intense transformation ('v5'), with performance rank indicated by proximity to the center, where 0 is the best. It shows that the performance of all algorithms deteriorates with increased transformation intensity, highlighted by the significant reordering of ranks and crossing of lines.}
\caption{Ranking of the performance of forecasting methods under magnitude warping transformation for the Tourism dataset, from original data (`orig`) to the most intense transformation (`v5`). Performance rank is indicated by proximity to the center, with 0 being the best. It shows that the performance of all algorithms deteriorates with increased transformation intensity, highlighted by the significant reordering of ranks and crossing of lines.}
\label{fig:ind_transf}
\end{figure}

We utilized four different transformations described in Section~\ref{chap:tranfs}. These transformations were applied to each dataset, and the parameters of each transformation were defined to increase linearly with a slope of 1.

The robustness of the following methods was analyzed:

\begin{itemize}
    \item \textbf{ETS + BU}: the ETS method was applied to the bottom time series, and then the naive Bottom-Up (BU)~\cite{fpp3} strategy was followed to aggregate the forecasts to the upper hierarchical levels.
    \item \textbf{ETS + MinT}: The ETS method was used as the base forecaster, but this time for all the time series (including the upper levels). To ensure the coherence of the forecasts, we used the Minimum Trace (MinT) reconciliation method proposed by~\cite{hyndman2018optimal}.
    \item \textbf{DeepAR + BU}: DeepAR produces probabilistic forecasts based on training an auto-regressive recurrent network model on related time series~\cite{deepAR_prob}. The hierarchy of the dataset is handled by representing it as multiple static categorical features to the model. We then use a naive BU reconciliation strategy.
    \item \textbf{TFT + BU}: Temporal Fusion Transformer (TFT) combines recurrent and transformer architectures for complex time series forecasting. TFT uses recurrent layers for local processing and interpretable self-attention layers for long-term dependencies~\cite{lim2020temporal}. We also use the naive BU reconciliation strategy after fitting the bottom series with TFT.
    \item \textbf{GPHF}: HTS forecasting model using Gaussian Processes~\cite{gphf}. No reconciliation strategy is needed with this approach as the model already takes into account the hierarchical structure of the data.
\end{itemize}

\subsection{Results and Discussion}

\begin{table*}[t]
\small
\centering
\begin{tabular}{llrrrrrrrrr}
\toprule
Dataset & Method & Bottom & Purpose & Region & State & Zone & Top \\
\hline
Tourism
& ETS + BU    & 0.86 & 1.08 & \textbf{0.84} & 1.06 & 0.95 & 1.50 \\
& ETS + MinT  & 0.88 & 1.03 & 0.85 & 1.04 & \textbf{0.93} & 1.43 \\
& DeepAR + BU     & 0.87 & 1.24 & \textbf{0.84} & 1.17 & 0.94 & 1.78 \\
& TFT + BU        & \textbf{0.82} & 1.74 & 0.87 & 1.47 & 1.08 & 2.59 \\
& GPHF        & 0.95 & \textbf{1.02} & 0.89 & \textbf{1.03} & 0.96 & \textbf{1.12} \\
\hline
Dataset & Method & Bottom   & Category & Depart. & Item   & State  & Store  & Top    \\
\hline
 M5
& ETS + BU        & 0.82 & 0.61 & 0.71   & 0.75 & 0.75 & 0.80 & 0.73 \\
& ETS + MinT      & 0.84 & \textbf{0.59} & 0.71   & 0.76 & \textbf{0.77} & 0.80 & 0.73 \\
& DeepAR + BU     & \textbf{0.75} & 0.71 & 0.77   & \textbf{0.70} & 0.95 & 0.92 & 1.06 \\
& TFT + BU        & \textbf{0.75} & 0.89 & 0.80   & \textbf{0.70} & 1.35 & 1.14 & 1.55 \\
& GPHF            & 0.90 & 0.66 & \textbf{0.70}   & 0.84 & 0.63 & \textbf{0.79} & \textbf{0.60} \\
\hline
Dataset & Method & Bottom & Crime & Beat & Street & Zip & Top \\
\hline
Police
& ETS + BU    & 0.98 & \textbf{0.82} & \textbf{0.84} & 0.77 & 0.83 & 0.85 \\
& ETS + MinT  & 0.95 & 0.85 & 0.83 & 0.84 & \textbf{0.82} & 1.34 \\
& DeepAR + BU & 1.03 & 0.85 & 0.88 & 0.77 & 0.86 & 0.92 \\
& TFT + BU    & \textbf{0.65} & 1.64 & 0.95 & 2.01 & 1.00 & 7.09 \\
& GPHF        & 0.99 & 0.87 & \textbf{0.84} & \textbf{0.76} & 0.85 & \textbf{0.77} \\
\hline
 \bottomrule
\end{tabular}
\caption{Results (MASE) for the original datasets considered in the experiments. \textbf{Bold} values represent the lowest error across models. The errors are calculated for each group, for bottom and top-level series.}
\label{table:original_res}
\end{table*}

We start by addressing \textbf{Q1} by evaluating the impact of our proposed transformations (Section~\ref{chap:tranfs}) in the distance distributions between the generated and original datasets. Remember that HTS algorithms rely on the dependencies between time series to improve the univariate estimates. Thus, we are interested in measuring how these relations are disrupted when applying each transformation controlled by its magnitude. Looking at Figure~\ref{fig:ridge}, we are systematically generating rich variations of the original datasets. Also, the behavior is consistent across all datasets.

If we look closer to Figure~\ref{fig:ridge}, for every transformation, the effect is limited and increases with the magnitude of the transformation, as expected. The transformation with the largest effect on the distance is magnitude warping, potentially having a greater impact on the performance of the algorithms that rely on the relations between the series. The jittering transformation reduces the spread of the distribution, i.e., it pushes the number of series that are very similar or dissimilar to each other to be closer to the original mean distance. Finally, as expected, the time warping transformation does not produce a relevant impact in the distance. We are using DTW to measure the distance, which already accounts for time-based warping effects. In this case, we are interested in the fact that although the distance does not change, there are warping effects in the time dimension that could impact performance.


Regarding \textbf{Q2}, Figure~\ref{fig:results} suggests that predictive performance varies significantly by transformation applied. Also, and answering \textbf{Q3}, increasing the magnitude of transformations has not only different effects but, in some cases, opposite ones. Figure~\ref{fig:results} lays out a comparative analysis of different HTS forecasting models (columns) in terms of their sensitivity to transformations (rows) in the time series data. We control by different levels of the hierarchy within the Tourism dataset — such as `bottom`, `purpose`, `region`, and so on. Note that as the x-axis increases, the magnitude of the transformation applied also increases.
Thus, a flatter line would suggest that a model is less sensitive to that particular transformation, maintaining a consistent performance despite the increasing intensity of data alteration. As anticipated, magnitude warping is the most disruptive transformation, and its impact increases with the magnitude. For all models, we can see that there is a positive slope, i.e., consistently worse results when the parameters of the transformation increase. Interestingly, for jitter, scaling, and time warping, the performance of all models stays flat or actually improves as we increase the magnitude of the transformation (especially for jitter). Comparing the robustness of the different models, we observe that GPHF is the model that shows more consistent behavior, which means less impact in terms of performance.

\begin{figure*}[t]
\centering
\includegraphics[width=\textwidth]{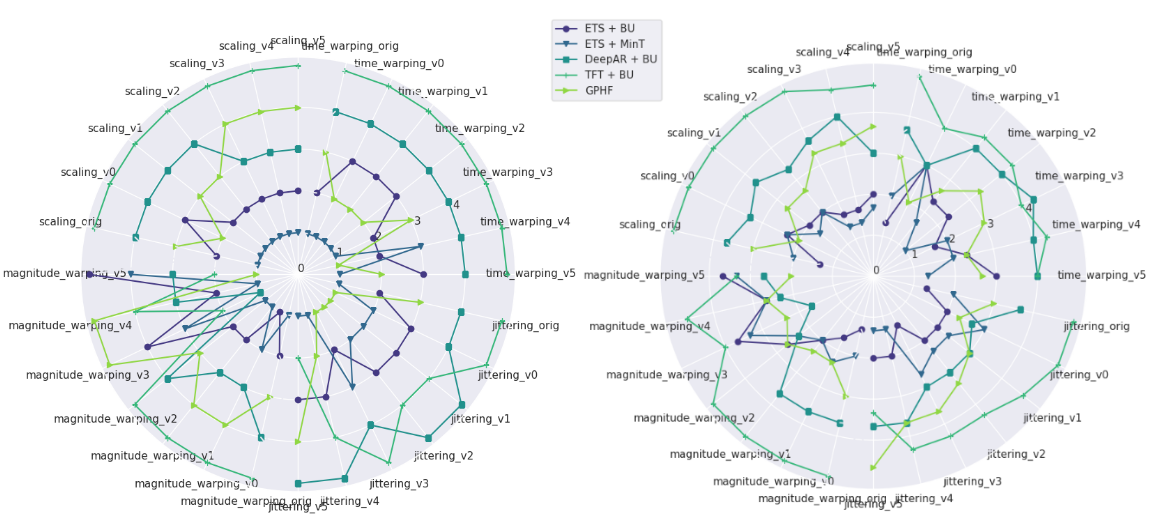}
\Description{The left chart shows the performance of forecasting algorithms against multiple transformations for the Tourism dataset. The right chart averages the performance ranks of forecasting algorithms for all datasets, providing a comparative overview of algorithm robustness across different transformations.}
\caption{The chart on the left shows the performance of forecasting algorithms against multiple transformations for the Tourism dataset. The chart on the right averages the performance ranks of forecasting algorithms for all datasets.}
\label{fig:ranks}
\end{figure*}

Regarding \textbf{Q4}, Figure~\ref{fig:ind_transf} and the left radar chart on Figure~\ref{fig:ranks} help us compare the robustness of the different models. The classical methods yield better results than the more complex counterparts in most cases. The only exception is when the transformation is too disruptive, such as magnitude warping, where we cannot spot any meaningful difference between the predictive performance of the different models.
The radar chart in Figure~\ref{fig:ind_transf} provides a visualization of the ranking of different forecasting models in handling the most disruptive transformation — magnitude warping — across various magnitudes of this transformation. Each axis represents a different set of parameters of the magnitude warping transformation, starting from the original data (`orig`) and progressing through increasingly intense transformations (`v0` through `v5`). The distance from the center indicates the rank of the method performance, with a rank of 1 being the closest to the center (best performance) and higher ranks being further away (worse performance).
The extensive crossing of lines as the magnitude increases shows the disruptive effect of the transformation. Thus, we cannot say that there is a model that would be more robust to this transformation when applied to this dataset in particular.
We then expanded the visualization to all transformations to have a global perspective on the robustness of the models (see the left radar chart on Figure~\ref{fig:ranks}). ETS + MinT method appears to be the most stable model across most transformations (except for magnitude warping, as already discussed). It also seems to suggest that using hierarchical information brings value to the model performance since ETS-BU is consistently worse than ETS + MinT. Most interestingly, classical methods seem to be the most robust when applied to this dataset.


When answering \textbf{Q5}, we start by applying a standard methodology to estimate the forecasting accuracy, which is often used to test algorithms (e.g., to compare a newly proposed algorithm with existing ones). In this analysis, researchers compare the predictive performance of algorithms on a small number of datasets. We reproduce what this analysis would look like, as shown in Table~\ref{table:original_res}. We can see that the results are somewhat consistent for the Tourism and Police datasets. If M5 was not considered, this could lead us to generalize on unreliable information. The main outcome is that we get mixed results, which are hard to generalize. Furthermore, the metrics presented in the table do not assess the robustness of the different algorithms.
 
Still, there are some insights that we can extract from this type of standard comparison. First, and contrary to our expectations, the ETS model does not show meaningful differences when using the naive reconciliation strategy or MinT. The global deep learning models performed well on the bottom series, especially TFT, but not on the aggregate-level ones. The GPHF model consistently produces good predictions for aggregated levels, but it is less accurate on the bottom level ones. Once again, can we generalize these results, or are they the result of our experimental setup?

After using RHiOTS across all datasets and averaging the results, we can more confidently answer \textbf{Q6} and \textbf{Q7} for all the algorithms. We don't see relevant differences between ETS-BU and ETS-MinT, and thus, using hierarchical information does not seem to produce increased robustness. On the predictive performance of algorithms, we consistently observe simpler algorithms (classical ones) yielding better results across transformations than more complex ones.
The radar chart on the right in Figure~\ref{fig:ranks} presents the average ranks of various forecasting algorithms across all datasets and controlled by transformation. It serves as a generalization of the chart on the left of the same figure. The chart shows that deep learning algorithms employing BU strategies, such as DeepAR and TFT, often underperform other approaches. The only exception is when faced with high-intensity magnitude warping. This is expected, as complex algorithms are better equipped to handle such behavior. The underperformance of these algorithms is not simply due to their naive reconciliation strategies, as evidenced by the resilience of the ETS + BU algorithm. GPHF ranks between deep learning algorithms and classical ones. Classical approaches, ETS + BU and ETS + MinT yield the best robustness, making them an appealing choice for use as a baseline when developing any new HTS algorithm.

\section{Conclusions and Future Work}
\label{chap:concl}

We propose RHiOTS, a novel framework for generating semi-synthetic time series datasets with controlled dependencies. We demonstrate its effectiveness in evaluating the performance of HTS models and algorithms under various conditions. Our empirical study using RHiOTS confirms that model and algorithm performance varies depending on the perturbations applied to the data and provides insights into the expected effects based on the type of perturbation.

First, we show that RHiOTS creates rich variations of the original datasets regarding the correlations between time series. When evaluating models, the predictive performance varies substantially based on the transformation applied. As we increase the magnitude, performance degrades quickly in transformations like magnitude warping, while it improves in cases such as jitter. When evaluating algorithms, we start by applying the conventional benchmark analysis of comparing predictive performance across three datasets. As the results were inconclusive, we extended the approach to use RHiOTS. One takeaway is that we do not see meaningful differences in robustness when applying specific reconciliation methods, such as MinT. The second is that classical algorithms are more robust than more complex deep learning counterparts. Deep learning algorithms only showed more robustness when the transformation was highly disruptive, such as high-intensity magnitude warping.
The aggregated findings from our visualizations and analyses suggest that if a single model must be chosen without prior knowledge of potential distortions in a dataset, the ETS model stands out as the most robust option.

Future research can focus on developing finer transformation controls, enabling a more targeted analysis of their impact on performance. Meta-learning techniques can help more effectively identify relationships between transformation parameters and model performance. RHiOTS, combined with these research directions, will help build a more comprehensive evaluation framework for hierarchical time series forecasting models.

\subsubsection*{Acknowledgements}

This work was partially funded by projects AISym4Med (101095387) supported by Horizon Europe Cluster 1: Health,  ConnectedHealth (n.º 46858), supported by Competitiveness and Internationalisation Operational Programme (POCI) and Lisbon Regional Operational Programme (LISBOA 2020), under the PORTUGAL 2020 Partnership Agreement, through the European Regional Development Fund (ERDF) and NextGenAI - Center for Responsible AI (C645008882-00000055), supported by IAPMEI,  and also by FCT plurianual funding for 2020-2023 of LIACC (UIDB/00027/2020\_UIDP/00027/2020)

\balance

\bibliographystyle{ACM-Reference-Format}
\bibliography{bib}

\end{document}